\documentclass[11pt,a4paper]{article}
\usepackage[hyperref]{naacl}
\usepackage{times}
\usepackage{url}
\usepackage{amsmath}
\usepackage{amssymb}
\usepackage{mathtools}
\usepackage{bm}
\usepackage{latexsym}
\usepackage{algorithm}
\usepackage[noend]{algpseudocode}
\usepackage{etoolbox}
\usepackage{float}
\usepackage{subfig}
\usepackage{multirow}
\usepackage{comment}
\usepackage{arydshln}
\usepackage{hyperref}
\aclfinalcopy 

\title{Relation Extraction using Explicit Context Conditioning}
\author{Gaurav Singh\thanks{~~G. Singh was an intern at Amazon at the time of work}\\University College London\\{g.singh@cs.ucl.ac.uk} \And Parminder Bhatia\\Amazon \\{parmib@amazon.com} }
\date{}

\PassOptionsToPackage{noend}{algpseudocode}
\usepackage{algpseudocode}

\errorcontextlines\maxdimen

\makeatletter

\begin{document}
\maketitle
\begin{abstract}
Relation Extraction (RE) aims to label relations between groups of marked entities in raw text. Most current RE models learn context-aware representations of the target entities that are then used to establish relation between them. This works well for intra-sentence RE and we call them first-order relations. However, this methodology can sometimes fail to capture complex and long dependencies. To address this, we hypothesize that at times two target entities can be explicitly connected via a context token. We refer to such indirect relations as second-order relations and describe an efficient implementation for computing them. These second-order relation scores are then combined with first-order relation scores. Our empirical results show that the proposed method leads to state-of-the-art performance over two biomedical datasets. 
\end{abstract}
\section{Introduction}



There are wide applications for Information Extraction in general \cite{jin2018improving} and Relation Extraction (RE) in particular, one reason why relation extraction continues to be an active area of  research \cite{bach2007survey,kambhatla2004combining,kumar2017survey}.  Traditionally, a standard RE model would start with entity recognition and then pass the extracted entities as inputs to a separate relation extraction model, which meant that the errors in entity recognition were propagated to RE. This problem was addressed by end-to-end models \cite{miwa2016end,zheng2017joint,adel2017global, bhatia2018end} that jointly learn both NER and RE. 


Generally, these models consist of an encoder followed by a relationship classification (RC) unit  \cite{verga2018simultaneously,christopoulou2018walk,su2017global}. The encoder provides context-aware vector representations for both target entities, which are then merged or concatenated before being passed to the relation classification unit, where a two layered neural network or multi-layered perceptron classifies the pair into different relation types.

Such RE models rely on the encoder to learn `perfect' context-aware entity representations that can capture complex dependencies in the text. This works well for intra-sentence relation extraction i.e. the task of extracting relation from entities contained in a sentence  \cite{christopoulou2018walk,su2017global}. As these entities are closer together, the encoder can more easily establish connection based on the language used in the sentence. Additionally, these intra-sentence RE models can use linguistic/syntactical features  for an improved performance e.g. shortest dependency path. 

\begin{figure}
    \centering
    \includegraphics[scale=0.4]{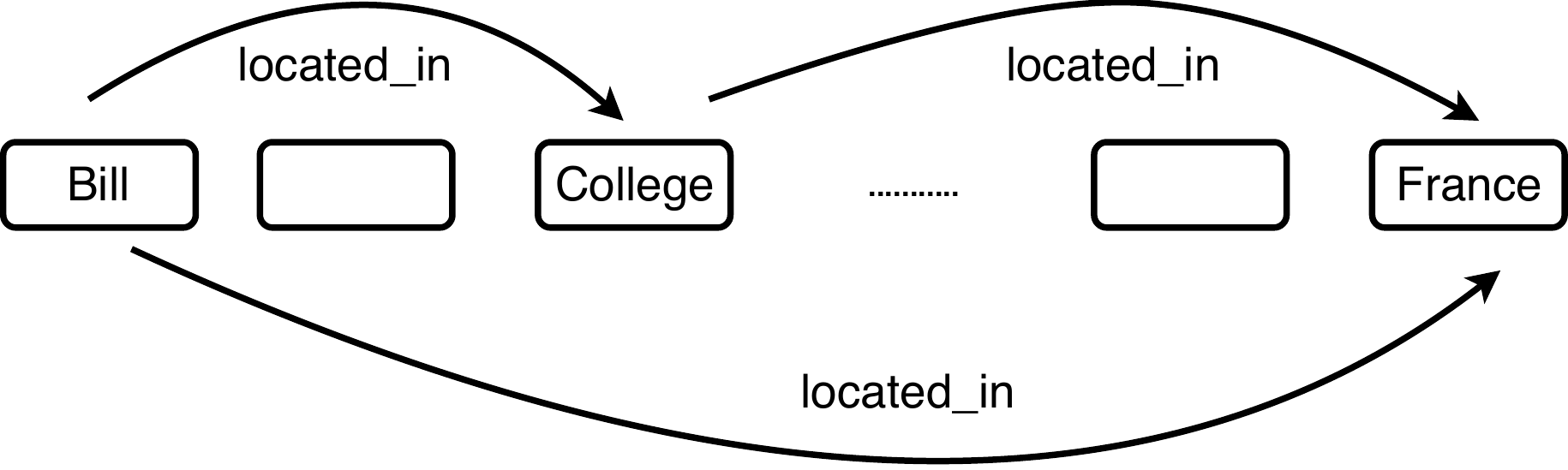}
    \caption{Pictorial representation of a second-order relation between two entities (\textit{Bill} \& \textit{France}) connected by a context token (\textit{College}).}
    \label{fig:hor}
     \vspace{-1.5em}
\end{figure}

Unfortunately, success in intra-sentence RE has not been replicated for cross-sentence RE. As an example, a recent RE method called BRAN \cite{verga2018simultaneously} proposed to use encoder of Transformer \cite{vaswani2017attention} for obtaining token representations and then used these representations for RE.  However, our analysis revealed that it wrongly marks many cross-sentence relations as negative, especially when the two target entities were connected by a string of logic spanning over multiple sentences. This showed that reliance on the encoder alone to learn these complex dependencies does not work well.


In this work we address this issue of over-reliance on the encoder. We propose a model based on the hypothesis that two target entities, whether intra-sentence or cross-sentence, could also be explicitly connected  via a third context token (Figure \ref{fig:hor}).  More specifically, we find a token in the text that is most related to both target entities, and compute the score for relation between the two target entities as the summation of their relation scores with this token. We refer to these relations as second-order relations. At the end, we combine these second-order scores with first-order scores derived from a traditional RE model, and achieve state-of-the-art performance over two biomedical datasets. 
To summarize the contribution of this work:
\begin{enumerate}
    \item We propose using second-order relation scores for improved relation extraction.
    \item We describe an efficient algorithm to obtain  second-order relation scores. 
\end{enumerate}



\section{Background}
In this section we describe the encoder and relation classification unit of a SOTA RE model called BRAN \cite{verga2018simultaneously}.  This model computes relation scores between two entities directly from their representations, therefore we refer to these as first-order relation scores.  
\subsection{Transformer Encoder}
\label{sec:encoder}
BRAN uses a variant of \textit{Transformer} \cite{vaswani2017attention} encoder to generate token representations.

The encoder contains repeating blocks and each such block consists of two sublayers:  multi-head self-attention layer followed by position-wise convolutional feedforward layer. There are residual connections and layer normalization \cite{ba2016layer} after each sublayer. The only difference from a standard transformer-encoder is the presence of a convolution layer of kernel width 5 between two consecutive convolution layers of kernels width 1 in the feedforward sublayer. It takes as input word embeddings that are added with positional embeddings \cite{gehring2017convolutional}.

\subsection{First-Order Relations}
The relation classification unit takes as input token representations from the described encoder. These are then passed through two MLPs to generate head/tail representation $e_i^{head}/e_i^{tail}$ for each token corresponding to whether it serves the first (head) or second (tail) position in the relation. 
\begin{equation}
\label{eq:head}
e_i^{head} = W_{head_2}(ReLU(W_{head_1} b_{i}))
\end{equation}
\vspace{-1em}
\begin{equation}
\label{eq:tail}
e_i^{tail} = W_{tail_2}(ReLU(W_{tail_1}b_{i}))
\end{equation}
where $b_i$ is the representation of the $i_{th}$ token generated by the encoder.

These are then combined with a bi-affine transformation operator to compute a $N\times R\times N$ tensor $A$ of pairwise affinity scores for every token pair and all relation types, scoring all  triplets of the form $(head, relation, tail)$:  
\begin{equation}
   A_{irj} = (e_i^{head}L) e_j^{tail},
\end{equation}
where $L$ is a learned tensor of dimension $d\times R \times d$ to map pairs of tokens to scores over each of the $R$ relation types and $d$ is the dimension of head/tail representations. Going forward we will drop the subscript $r$ for clarity. 

The contributions from different mention pairs are then aggregated to give us \textbf{first-order} relation scores. This aggregation is done using $LogSumExp$, which is a smooth approximation of $max$ that prevents sparse gradients:
\begin{equation}
    \text{scores}^{(1)}(p^{head},p^{tail}) = \log\sum_{\mathclap{\substack{i \in P^{head}\\ j \in P^{tail}}}} \exp(A_{ij}),
    \label{eq:foscore}
\end{equation}
where $P^{head} (P^{tail})$ contains mention indices for head (tail) entity.

\section{Proposed Second-Order Relations}
In this section we describe in detail our proposed method to obtain second-order relation scores. 

We use the encoder described in Sec \ref{sec:encoder} for getting token representations. These token representations are then passed through two MLPs (as in previous section), which generate head/tail representations for each token corresponding to whether it serves the first or the second position in the relation. We used a separate set of these head/tail MLPs for second-order scores than the ones used for getting first-order scores. This was motivated by the need for representations focused on establishing relations with context tokens, as opposed to first-order relations (described in previous section) that attempt to directly connect two target entities.  


The head and tail representations are then combined with a $d\times R\times d$ bilinear transformation tensor $M$ to get a $N\times R \times N$ tensor $B$ of intermediate pairwise scores.
\begin{figure}
    \centering
    \includegraphics[scale=0.6]{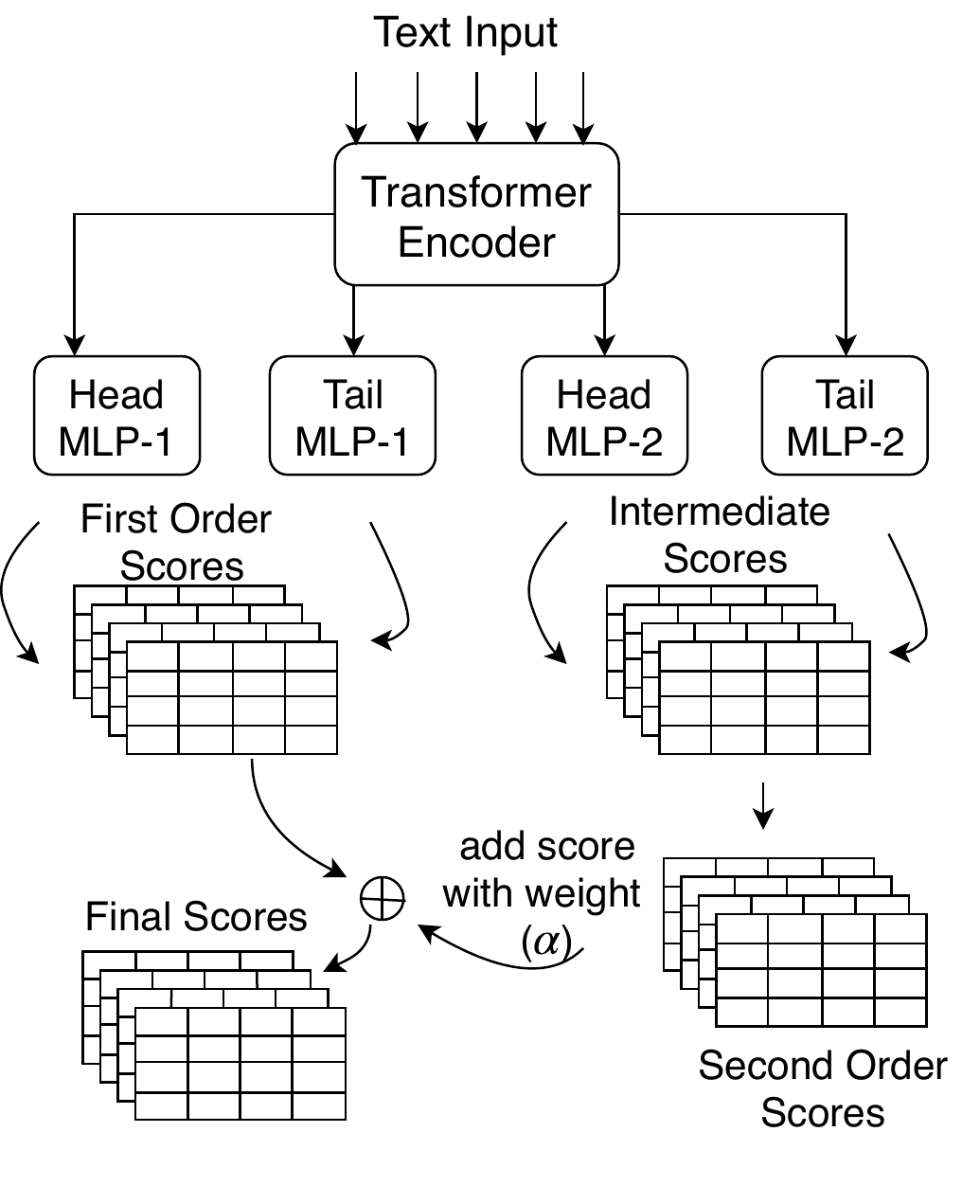}
    \vspace{-1em}
    \caption{Schematic of the model architecture. 
    }
    \label{fig:model_arch}
    \vspace{-1em}
\end{figure}
\begin{equation}
B_{ij} = (e_i^{head}M) e_j^{tail}
\label{eq:ips}
\end{equation}

After that we arbitrarily define the scores between tokens $i$ and $j$ when \textit{conditioned} on a context token $k$ as the sum of the scores of relations $(i,k)$ and $(k,j)$. 
\begin{equation}
C(i,j|k) = B_{ik} + B_{kj}
\end{equation}
These context-conditioned scores are computed for every triplet of the form $(i,j,k)$. 

\textbf{Second-order} relation scores are then derived by aggregating over all context tokens and mention pairs using $LogSumExp$.  
\begin{multline}
\text{scores}^{(2)}(p^{head},p^{tail}) 
= \log\sum_{\mathclap{\substack{k\\ i\in P^{head} \\ j\in P^{tail}}}} \exp(C(i,j|k))
\label{eq:soscore}
\end{multline}
Here $LogSumExp$ ensures that one specific mention pair connected via one specific context token is responsible for the relation.
This is equivalent to max-pooling over all context tokens that could potentially connect the two target entities, which reduces over-fitting by removing contributions from noisy associations of the target entities with random tokens e.g. stopwords. 

It is important to mention that a naive implementation of this would require $O(N^3)$ space to store context-conditioned scores between all pairs of token i.e. $C(i,j|k)$. To address this, we describe an efficient method in Section \ref{sec:EI} that avoids explicitly storing these.

At the end, the final score for relation between two entities is given as a weighted sum of first (eq. \ref{eq:foscore}) and second (eq. \ref{eq:soscore}) order scores. 
\begin{equation}
\begin{split}
\text{scores}(p^{head},p^{tail}) &= ~\text{scores}^{(1)}(p^{head},p^{tail})\\
& + \alpha * \text{scores}^{(2)}(p^{head},p^{tail})
\end{split}
\label{eq:final}
\end{equation}
where $\alpha$ is a hyper-parameter denoting the weight of second-order relation scores.\\

\noindent
\textbf{Entity Recognition}. We do entity recognition alongside relation extraction, as the transfer of knowledge between the two tasks has been shown to improve relation extraction performance  \cite{verga2018simultaneously,miwa2016end}. For this we feed encoder output $b_i$ to a linear classifier $W_{er}$ that predicts  scores for each entity type.
\begin{equation}
    d_i = W_{er}(b_i)
\end{equation}

\subsection{Efficient Implementation}
\label{sec:EI}
The problem lies in storing score for every intermediate relation of the form $C(i,j|k)$, as that would require space of the order $O(N^3)$. Here we describe a space-time efficient method to compute final second-order relation scores.

The intermediate scores (eq. \ref{eq:ips}) are a tensor of dimension $b \times N \times R \times N$ comprising of pairwise scores for $b$ batches. We create two tensors out of these intermediate scores, namely $T_1$ and $T_2$. $T_1$ computes the exponential of indices ($\{b, i \in P^{head}, j \in \mathcal{C}, R\}$) corresponding to pairwise scores between head entity and all the context tokens ($\mathcal{C}$ i.e., all the tokens except the two target entities), and sets other indices to $0$. Similarly, $T_2$  computes exponential of indices ($\{ b, i \in P^{tail}, j \in \mathcal{C},  R\}$) corresponding to pairwise scores between tail entity and context tokens, setting all other indices to $0$. To get the context conditioned scores one needs to compute the batch product of $R$ two dimensional slices of size $N\times N$ from $T_1$ and $T_2$ along the dimension of context, but this would be sequential in $R$.
Instead we can permute $T_1$ and $T_2$ to $b \times R \times N \times N$ followed by reshaping to $bR\times N \times N$ and perform a batch matrix multiplication along the context dimension to get $bR\times N \times N$.  Afterwards, we can sum along the last two dimensions to get a tensor of size $bR$. Finally, we can take the log succeeded by reshaping to $b \times R$ to obtain second-order scores.  


\section{Experimentation}

\begin{table}[t]
\begin{center}
\begin{tabular}{ l  l | c  c  c }
Data & Model & Pr & Re & F1  \\\hline
\multirow{2}{*}{DCN} & BRAN & 0.614 & 0.850 & 0.712  \\ 
                         & + SOR & \textbf{0.643} & \textbf{0.879} & \textbf{0.734}  \\ \hline
\multirow{3}{*}{i2b2}    & HDLA   & 0.378 & \textbf{0.422} & 0.388 \\
                         & BRAN & 0.396 & 0.403 & 0.395  \\
                         & + SOR & \textbf{0.424} & 0.419 & \textbf{0.407} \\ \hline
                         
\multirow{2}{*}{CDR}   & BRAN & 0.552 & 0.701 & 0.618\\
                         & + SOR & 0.552 & 0.701 & 0.618 \\
                         
\end{tabular}
\end{center}
\vspace{-1em}
\caption{The performance of proposed model using second-order relations. BRAN is the model used in \protect\cite{verga2018simultaneously} and +SOR is our proposed model with second-order relations. Results for HDLA are quoted from \protect\citet*{chikka2018hybrid}. Results on CDR are identical for both BRAN and our proposed model as $\alpha$ was set to $0$ after tuning over the dev set at which point our model is the same as BRAN.  All the metrics are macro in nature.}
\label{tab:results}
\vspace{-1.5em}
\end{table}

%
\subsection{Datasets}
We have used three datasets in this work, i2b2 2010 challenge \cite{uzuner20112010} dataset, a de-identified clinical notes dataset and a chemical-disease relations dataset known as BioCreative V (CDR) \cite{li2016biocreative, wei2016assessing}. 

\textit{First} is a publicly available subset of the dataset used for the i2b2 2010 challenge. It consists of documents describing relations between different diseases and treatments.  Out of the 426 documents available publicly, 10\% are used each for both dev and test and the rest for training.  There are 3244/409 relations in train/test set and 6 pre-defined relations types including one negative relation  e.g. TrCP (\textbf{Tr}eatment Causes \textbf{Pr}oblem), TrIP (Tr Improves Pr), TrWP (Tr Worsens Pr). We have used the exact same dataset as Chikka \textit{et al.} \cite{chikka2018hybrid}.  


\textit{Second} is a dataset of 4200 de-identified clinical notes (DCN), with vocabulary size of 50K. It contains approximately 170K relations in the train set and 50K each in dev/test set. There are 7 pre-defined relation types including one negative relation type. These are mostly between medication name and other entities e.g. ``\textit{paracetamol \textbf{every} day}",``\textit{aspirin \textbf{with\_dosage} 100mg}". The frequency of different relations in this dataset is fairly balanced. 

\textit{Third} is a widely used and publicly available dataset called CDR \cite{li2016biocreative, wei2016assessing}. It was derived from Comparative Toxicogenomics Database (CTD) and contains documents describing the effect of chemicals (drugs) on diseases. There are only two relation types between any two target entities i.e. positive/negative and these relations are annotated at the document level. It consists of 1500 documents that are divided equally between train/dev/test sets. There are 1038/1012/1066 positive and 4280/4136/4270 negative relations in train/dev/test sets respectively. We performed the same preprocessing as done in BRAN \cite{verga2018simultaneously}.

\vspace{-0.5em}
\subsection {Experimental Settings}

We jointly solve for NER and RE tasks using cross-entropy loss. During training we alternate between mini-batches derived from each task. We fix the learn rate to $0.0005$ and  clip gradient for both tasks at  $5.0$. For training, we used adams optimizer with $\beta=(\beta_1,\beta_2)=(0.1,0.9)$. We tune over the weight of second-order relations denoted by $\alpha$ to get $\alpha=0.2$ for DCN/i2b2 and $\alpha=0.0$ for CDR dataset. 

Our final network had two encoder layers, with 8 attention heads in each multi-head attention sublayer and 256 filters for convolution layers in position-wise feedforward sublayer. We used dropout with probability $0.3$ after: embedding layer, head/tail MLPs, output of each encoder sublayer. We also used a word dropout with probability $0.15$ before the embedding layer.


\subsection{Results}
To show the benefits of using second-order relations we compared our model's performance to BRAN. The two models are different in the weighted addition of second-order relation scores. We tune over this weight parameter on the dev set and observed an improvement in MacroF1 score from $0.712$ to $0.734$ over DCN data and from $0.395$ to $0.407$ over i2b2 data. For further comparison a recently published model called HDLA \cite{chikka2018hybrid} reported a macro-F1 score of $0.388$ on the same i2b2 dataset. 
It should be mentioned that HDLA used syntactic parsers for feature extraction but we do not use any such external tools. 

In the case of CDR dataset we obtained $\alpha=0$ after tuning, which means that the proposed model converged to BRAN and the results were identical for the two models.  These results are summarized in Table \ref{tab:results}.

\subsection{Ablation Study} 
We experimented with different ablations of BRAN and 
noticed an improvement in results for DCN dataset upon removing multi-head self-attention layer. Also, our qualitative analysis showed that relations between distant entities were often wrongly marked negative. We attribute these errors to the token representations generated by the encoder. To this effect, our experiments showed that incorporating relative position \cite{DBLP:conf/naacl/ShawUV18} information in the encoder to improve token representations does not lead to superior RE.  Separately, we observed that the proposed method improved results when using a standard CNN encoder as well.   


\section{Conclusions and Future Work}
We proposed a method that uses second-order relation scores to capture long dependencies for improved RE.  These relations are derived by explicitly connecting two target entities via a context token. These second-order relations (SORs) are then combined with traditional relation extraction models, leading to state-of-the-art performance over two biomedical datasets. We also describe an efficient implementation for obtaining these SORs.


Despite restricting ourselves to SORs, it should be noted that the proposed method can be generalized to third and fourth order relations. We conjecture that these may serve well for cross-sentence relation extraction in long pieces of texts.  Also, we only considered one relation type between each entity and bridge token but it is possible, and very likely that two different relation types may lead to a third relation type. We will explore both these aspects in future work.

\section{Acknowledgement}
We would like to thank Busra Celikkaya and Mohammed Khalilia of Amazon, Zahra Sabetsarvestani and Sebastian Riedel of University College London and the anonymous reviewers of NAACL for their valuable feedback on the paper.
\bibliography{naaclhlt2018}

\begin{thebibliography}{19}
\expandafter\ifx\csname natexlab\endcsname\relax\def\natexlab#1{#1}\fi

\bibitem[{Adel and Sch{\"{u}}tze(2017)}]{adel2017global}
Heike Adel and Hinrich Sch{\"{u}}tze. 2017.
\newblock Global normalization of convolutional neural networks for joint
  entity and relation classification.
\newblock In \emph{Proceedings of the 2017 Conference on Empirical Methods in
  Natural Language Processing, {EMNLP} 2017}.

\bibitem[{Ba et~al.(2016)Ba, Kiros, and Hinton}]{ba2016layer}
Jimmy~Lei Ba, Jamie~Ryan Kiros, and Geoffrey~E Hinton. 2016.
\newblock Layer normalization.
\newblock \emph{arXiv preprint arXiv:1607.06450}.

\bibitem[{Bach and Badaskar(2007)}]{bach2007survey}
Nguyen Bach and Sameer Badaskar. 2007.
\newblock A survey on relation extraction.
\newblock \emph{Language Technologies Institute, Carnegie Mellon University}.

\bibitem[{Bhatia et~al.(2018)Bhatia, Celikkaya, and Khalilia}]{bhatia2018end}
Parminder Bhatia, Busra Celikkaya, and Mohammed Khalilia. 2018.
\newblock End-to-end joint entity extraction and negation detection for
  clinical text.
\newblock \emph{arXiv preprint arXiv:1812.05270}.

\bibitem[{Chikka and Karlapalem(2018)}]{chikka2018hybrid}
Veera~Raghavendra Chikka and Kamalakar Karlapalem. 2018.
\newblock A hybrid deep learning approach for medical relation extraction.
\newblock \emph{arXiv preprint arXiv:1806.11189}.

\bibitem[{Christopoulou et~al.(2018)Christopoulou, Miwa, and
  Ananiadou}]{christopoulou2018walk}
Fenia Christopoulou, Makoto Miwa, and Sophia Ananiadou. 2018.
\newblock A walk-based model on entity graphs for relation extraction.
\newblock In \emph{Proceedings of the 56th Annual Meeting of the Association
  for Computational Linguistics (Volume 2: Short Papers)}, volume~2, pages
  81--88.

\bibitem[{Gehring et~al.(2017)Gehring, Auli, Grangier, Yarats, and
  Dauphin}]{gehring2017convolutional}
Jonas Gehring, Michael Auli, David Grangier, Denis Yarats, and Yann~N Dauphin.
  2017.
\newblock Convolutional sequence to sequence learning.
\newblock \emph{arXiv preprint arXiv:1705.03122}.

\bibitem[{Jin et~al.(2018)Jin, Bahadori, Colak, Bhatia, Celikkaya, Bhakta,
  Senthivel, Khalilia, Navarro, Zhang et~al.}]{jin2018improving}
Mengqi Jin, Mohammad~Taha Bahadori, Aaron Colak, Parminder Bhatia, Busra
  Celikkaya, Ram Bhakta, Selvan Senthivel, Mohammed Khalilia, Daniel Navarro,
  Borui Zhang, et~al. 2018.
\newblock Improving hospital mortality prediction with medical named entities
  and multimodal learning.
\newblock \emph{arXiv preprint arXiv:1811.12276}.

\bibitem[{Kambhatla(2004)}]{kambhatla2004combining}
Nanda Kambhatla. 2004.
\newblock Combining lexical, syntactic, and semantic features with maximum
  entropy models for extracting relations.
\newblock In \emph{Proceedings of the ACL 2004 on Interactive poster and
  demonstration sessions}, page~22. Association for Computational Linguistics.

\bibitem[{Kumar(2017)}]{kumar2017survey}
Shantanu Kumar. 2017.
\newblock A survey of deep learning methods for relation extraction.
\newblock \emph{arXiv preprint arXiv:1705.03645}.

\bibitem[{Li et~al.(2016)Li, Sun, Johnson, Sciaky, Wei, Leaman, Davis,
  Mattingly, Wiegers, and Lu}]{li2016biocreative}
Jiao Li, Yueping Sun, Robin~J Johnson, Daniela Sciaky, Chih-Hsuan Wei, Robert
  Leaman, Allan~Peter Davis, Carolyn~J Mattingly, Thomas~C Wiegers, and Zhiyong
  Lu. 2016.
\newblock Biocreative v cdr task corpus: a resource for chemical disease
  relation extraction.
\newblock \emph{Database}, 2016.

\bibitem[{Miwa and Bansal(2016)}]{miwa2016end}
Makoto Miwa and Mohit Bansal. 2016.
\newblock End-to-end relation extraction using lstms on sequences and tree
  structures.
\newblock In \emph{Proceedings of the Annual Meeting of the Association for
  Computational Linguistics, {ACL} 2016}.

\bibitem[{Shaw et~al.(2018)Shaw, Uszkoreit, and
  Vaswani}]{DBLP:conf/naacl/ShawUV18}
Peter Shaw, Jakob Uszkoreit, and Ashish Vaswani. 2018.
\newblock Self-attention with relative position representations.
\newblock In \emph{Proceedings of the 2018 Conference of the North American
  Chapter of the Association for Computational Linguistics: Human Language
  Technologies, NAACL-HLT}.

\bibitem[{Su et~al.(2018)Su, Liu, Yavuz, Gur, Sun, and Yan}]{su2017global}
Yu~Su, Honglei Liu, Semih Yavuz, Izzeddin Gur, Huan Sun, and Xifeng Yan. 2018.
\newblock Global relation embedding for relation extraction.
\newblock In \emph{Proceedings of the 2018 Conference of the North American
  Chapter of the Association for Computational Linguistics: Human Language
  Technologies, {NAACL-HLT} 2018}.

\bibitem[{Uzuner et~al.(2011)Uzuner, South, Shen, and DuVall}]{uzuner20112010}
{\"O}zlem Uzuner, Brett~R South, Shuying Shen, and Scott~L DuVall. 2011.
\newblock 2010 i2b2/va challenge on concepts, assertions, and relations in
  clinical text.
\newblock \emph{Journal of the American Medical Informatics Association},
  18(5).

\bibitem[{Vaswani et~al.(2017)Vaswani, Shazeer, Parmar, Uszkoreit, Jones,
  Gomez, Kaiser, and Polosukhin}]{vaswani2017attention}
Ashish Vaswani, Noam Shazeer, Niki Parmar, Jakob Uszkoreit, Llion Jones,
  Aidan~N Gomez, {\L}ukasz Kaiser, and Illia Polosukhin. 2017.
\newblock Attention is all you need.
\newblock In \emph{Advances in Neural Information Processing Systems}.

\bibitem[{Verga et~al.(2018)Verga, Strubell, and
  McCallum}]{verga2018simultaneously}
Patrick Verga, Emma Strubell, and Andrew McCallum. 2018.
\newblock Simultaneously self-attending to all mentions for full-abstract
  biological relation extraction.
\newblock In \emph{Proceedings of the 2018 Conference of the North American
  Chapter of the Association for Computational Linguistics: Human Language
  Technologies, {NAACL-HLT} 2018}.

\bibitem[{Wei et~al.(2016)Wei, Peng, Leaman, Davis, Mattingly, Li, Wiegers, and
  Lu}]{wei2016assessing}
Chih-Hsuan Wei, Yifan Peng, Robert Leaman, Allan~Peter Davis, Carolyn~J
  Mattingly, Jiao Li, Thomas~C Wiegers, and Zhiyong Lu. 2016.
\newblock Assessing the state of the art in biomedical relation extraction:
  overview of the biocreative v chemical-disease relation (cdr) task.
\newblock \emph{Database}, 2016.

\bibitem[{Zheng et~al.(2017)Zheng, Wang, Bao, Hao, Zhou, and
  Xu}]{zheng2017joint}
Suncong Zheng, Feng Wang, Hongyun Bao, Yuexing Hao, Peng Zhou, and Bo~Xu. 2017.
\newblock Joint extraction of entities and relations based on a novel tagging
  scheme.
\newblock In \emph{Proceedings of the Annual Meeting of the Association for
  Computational Linguistics, {ACL} 2017}.

\end{thebibliography}
\bibliographystyle{acl_natbib_nourl}

\end{document}